\documentclass[letterpaper]{IEEEtran}
\usepackage{latexsym,,amssymb,amsmath,graphicx,epsf,cite,bbm,float}
\usepackage{ifpdf}
\usepackage{epstopdf}

\usepackage{algorithm,algorithmic}
\usepackage{amsmath,amssymb,bm}
\usepackage{amsfonts,dsfont,color,bbm,subcaption}

\usepackage{amsmath,amsthm}
\usepackage{hyperref}

\newtheorem{theorem}{Theorem}

\newtheorem{lemma}[]{Lemma}

\newtheorem{remark}[]{Remark}

\newtheorem{definition}[]{Definition}

\newcommand{\be}{\begin{equation}}
\newcommand{\ee}{\end{equation}}
\newcommand{\bea}{\begin{eqnarray}}
\newcommand{\eea}{\end{eqnarray}}

\newcommand{\MB}{\left[\begin{array}}
\newcommand{\ME}{\end{array}\right]}

\newcommand{\ei}{\end{itemize}}
\newcommand{\bi}{\begin{itemize}}

\newcommand{\norm}[1]{\lVert#1\rVert}

\usepackage{amsmath,amsthm}

\DeclareMathOperator*{\argmin}{arg\,min}

\title{Efficient Minimax Optimal Global Optimization of Lipschitz Continuous Multivariate Functions}
\author{\IEEEauthorblockN{Kaan Gokcesu}, \IEEEauthorblockN{Hakan Gokcesu} }
\begin{document}
\maketitle

\begin{abstract}
	In this work, we propose an efficient minimax optimal global optimization algorithm for multivariate Lipschitz continuous functions. 
	To evaluate the performance of our approach, we utilize the average regret instead of the traditional simple regret, which, as we show, is not suitable for use in the multivariate non-convex optimization because of the inherent hardness of the problem itself. Since we study the average regret of the algorithm, our results directly imply a bound for the simple regret as well. 
	Instead of constructing lower bounding proxy functions, our method utilizes a predetermined query creation rule, which makes it computationally superior to the Piyavskii-Shubert variants.
	We show that our algorithm achieves an average regret bound of $O(L\sqrt{n}T^{-\frac{1}{n}})$ for the optimization of an $n$-dimensional $L$-Lipschitz continuous objective in a time horizon $T$, which we show to be minimax optimal.
\end{abstract}

\section{Introduction}
	
	In the problem of global optimization \cite{pinter1991global}, the goal is to minimize an objective function with as few evaluations as possible. It can be utilized in many applications such as hyper-parameter tuning in complex learning systems \cite{cesa_book, poor_book}. In these applications, because of the complexity of the system, the loss function in question does not generally possess properties that ease optimization (e.g., convexity). The sequential optimization of such unknown and possibly non-convex functions are called by many names, which includes:
	\begin{itemize}
		\item global optimization \cite{pinter1991global}
		\item derivative-free optimization \cite{rios2013derivative}
		\item black-box optimization \cite{jones1998efficient}. 
	\end{itemize}

	The problem of global optimization has attracted significant attention, especially in recent years, in a wide variety of research fields including non-convex optimization \cite{jain2017non, hansen1991number, basso1982iterative}, Bayesian optimization \cite{brochu2010tutorial}, convex optimization \cite{boyd2004convex,nesterov2003introductory,bubeck2015convex}, bandit optimization \cite{munos2014bandits}, stochastic optimization \cite{shalev2012online, spall2005introduction}. It can also be straightforwardly utilized in many applications such as distribution estimation \cite{gokcesu2018density,willems,gokcesu2018anomaly,coding2}, multi-armed bandits \cite{neyshabouri2018asymptotically,cesa-bianchi,gokcesu2018bandit}, control theory \cite{tnnls3}, signal processing \cite{ozkan}, game theory \cite{tnnls1}, prediction \cite{gokcesu2016prediction,singer}, decision theory \cite{tnnls4} and anomaly detection \cite{gokcesu2019outlier,gokcesu2016nested,gokcesu2017online}.
	
	In global optimization, when optimizing a function $f(\cdot)$, we only have access to its evaluations at certain query points.
	Although there exist many heuristics, we focus on the regularity based approaches \cite{bartlett2019simple,grill2015black}. Specifically, we focus on the more popular Lipschitz regularity, which was first utilized by Piyavskii for univariate Lipschitz continuous functions, where the approach is to create lower bounding proxy functions on the objective and to query their minimum \cite{piyavskii1972algorithm}.
	This method was also developed by Shubert independently in the same year \cite{shubert1972sequential}; hence, it has been generally called as the Piyavskii-Shubert algorithm.
	
	There have been many developments on this algorithm over the years to study its varying specific applications to functions with distinct Lipschitz regularity conditions \cite{basso1982iterative,schoen1982sequential,mayne1984outer,mladineo1986algorithm,shen1987interval,horst1987convergence,hansen1991number,hansen1995lipschitz,horst2013global}.
	In \cite{breiman1993deterministic}, a multivariate extension to the original algorithm was proposed that utilizes the Taylor expansion of the objective function.
	In \cite{baritompa1994accelerations}, an acceleration of this multivariate extension has been designed. 
	In \cite{sergeyev1998global}, the use of smooth auxiliaries have been considered to create the lower bounding proxy functions in the Piyavskii-Shubert algorithm. In \cite{ellaia2012modified}, a variant has been created for use in differentiable univariate functions.
	Similarly, another variant where the objective function is defined on a compact interval with a bounded second derivative is also proposed in \cite{brent2013algorithms}. Another work studies the univariate global optimization for the objective functions with generalized Lipschitz regularities \cite{gokcesu2021regret}.
	
	The performance analysis in regards to the algorithms for global optimization is done via the convergence to the functional evaluation of an optimizer of the objective $f(\cdot)$ \cite{danilin1971estimation}. Traditionally, the performance analysis has been done via the notion of simple regret, which measures the difference between the evaluation of the best query so far with the evaluation of an optimizer. It has been shown that Piyavskii-Shubert algorithm has a crude regret bound of the form $\tilde{r}_T = O(T^{-\frac{1}{n}})$ for a Lipschitz continuous $n$-dimensional objective \cite{mladineo1986algorithm}. 
	For univariate functions, it has been shown that a Piyavskii–Shubert variant can achieve a simple regret of $\epsilon$ in at most $O(\int_0^1(f(x)-f(x^*)+\epsilon)^{-1}dx)$ number of queries \cite{hansen1991number}, which improves upon the results of \cite{danilin1971estimation}. Similarly, the work in \cite{ellaia2012modified} improves upon the results of its predecessors \cite{danilin1971estimation,hansen1991number}. A variant of the Piyavskii-Shubert algorithm called LIPO has been proposed in \cite{malherbe2017global}, which achieves better upper bounds on the simple regret under stronger assumptions. Another work studies the simple regret of the Piyavskii-Shubert algorithm under noisy evaluations \cite{bouttier2020regret}.
	
	As opposed to the weaker simple regret, the cumulative regret bounds of the Piyavskii-Shubert variant algorithms have also been studied for univariate functions \cite{gokcesu2021regret}. While the Piyavskii-Shubert type methods have low cumulative regret bounds for Lipschitz continuous functions, the optimization of the lower bounding functions in the process of determining the queries can be costly. Another work \cite{gokcesu2022low} circumvents this issue by utilizing predetermined sampling sets and achieves similar cumulative regret bounds with Piyavskii-Shubert algorithms in univariate settings. However, in all these methods the objective function takes a single argument (i.e., univariate). To this end, we extend these approaches to the more general multivariate case (dimension $n>1$) and create an algorithm that achieves minimax optimal cumulative regret bound for Lipschitz continuous functions.
	
\section{Preliminaries}	
\subsection{Problem definition}\label{sec:problem}
In this section, we formalize the multivariate global optimization problem. We want to minimize a multivariate function $f(\cdot)$  such that
\begin{align}
	f(\cdot) : \Omega\rightarrow \Re,\label{eq:f}
\end{align}
i.e., it maps from a compact subset $\Omega\subset \Re^n$ to the set of real numbers $\Re$. We will mainly consider the case where $\Omega$ is the unit $n$-dimensional cube, i.e.,
\begin{align}
	\Omega\equiv [0,1]^n.
\end{align} 

Even though the objective $f(\cdot)$ is potentially non-convex (i.e., does not possess the desirable property of convexity), it is not an arbitrary function and commonly possesses some type of regularity. We consider the widely used Lipschitz continuity regularity.

\begin{definition}\label{def:condition}
	Let the objective function $f(\cdot)$ satisfy the following regularity:
	\begin{align*}
		|f(x)-f(y)|\leq L\norm{x-y},
	\end{align*}
	for any $x,y\in\Omega$, where
	\begin{itemize}
		\item $L>0$ is the Lipschitz continuity coefficient, 
		\item $\lVert\cdot\rVert$ is the Euclidean norm. 
	\end{itemize}
\end{definition}

We iteratively minimize $f(\cdot)$ by observing the evaluation $f(x_t)$ of a query $x_t$ and selecting the next query $x_{t+1}$ based upon the already evaluated queries $\{x_t\}_{t=1}^T$ and their evaluations $\{f(x_t)\}_{t=1}^T$, i.e.,
\begin{align}
	x_{t+1}=\Gamma(x_1,x_2,\ldots,x_{t},f(x_1),f(x_2),\ldots,f(x_{t})),
\end{align}
for some decision making system denoted by the function $\Gamma(\cdot)$. Example decision functions are implied by the popular Piyavskii–Shubert variants, where the next potential query is determined by constructing the lower bounding proxy function that crosses from the already queried points and their evaluations \cite{gokcesu2021regret}.

The goal of the learning process (and querying) is to determine the optimal points with as few evaluations as possible. From the perspective of the computational learning theory, this coincides with considering the objective function as a loss function and each evaluation 
\begin{align}
	f(x_t)\in\Re,
\end{align}
as the loss incurred when the query point 
\begin{align}
	x_t\in\Omega,
\end{align}
is selected. Since the function $f(\cdot)$ can have arbitrarily high evaluations, we use the notion of regret and compare the evaluations of our selections with an optimal evaluation for the performance analysis.
Let $x_*$ be a global minimizer of $f(\cdot)$, i.e.,
\begin{align}
	f(x_*) = \min_{x\in\Omega} f(x).\label{eq:min}
\end{align} 

Next, we show why the notion of average regret is more useful for the performance evaluation of global optimization algorithms in contrast to the simple regret.

\subsection{Simple Regret vs Average Regret}
For a time horizon $T$ (number of evaluations), there exists two schools of though for the regret analysis:
\begin{itemize}
	\item The first one is the traditional simple regret, where the queried points does not result in a loss and only the selected points have a loss incurred. After querying the points $\{x_t\}_{t=1}^T$, we select a point $\tilde{x}_T$, which will result in a loss. The simple regret is given by the difference to the minimum possible evaluation of $f(\cdot)$, hence,
	\begin{align}
		\tilde{r}_T=f(\tilde{x}_T)-\min_{x\in\Omega}f(x).
	\end{align}

	\item The second approach is the average regret analysis, where all the queries made $\{x_t\}_{t=1}^T$ contribute to the loss incurred. The average regret is given by the distance to the minimum possible evaluation of $f(\cdot)$, hence,
	\begin{align}
		r_T=\frac{1}{T}\sum_{t=1}^{T}f({x}_t)-\min_{x\in\Omega}f(x).
	\end{align}
\end{itemize}

While the simple regret analysis or convergence to the optimal evaluation may be useful in various problem scenarios, it is not meaningful for use in performance evaluation of an algorithm for global optimization problems because of the inherent hardness of the problem. To see this, consider the optimization of an $L$-Lipschitz continuous function (as in \autoref{def:condition}) over the set $[0,1]^n$. Let us split this $n$-dimensional unit cube into $n$-dimensional cubes of side lengths $2\epsilon$ for some small $\epsilon$. Let a rudimentary grid search algorithm sample the center of these small $n$-dimensional cubes, i.e., $\{\bar{x}_t\}_{t=1}^{T_\epsilon}$, where the number of samples $T_\epsilon$ is given by the following
\begin{align}
	T_\epsilon=O\left(\frac{1}{(2\epsilon)^n}\right).
\end{align}
The simple regret of this grid search is given by
\begin{align}
	\tilde{r}_{T_\epsilon}\leq& \min_{t\in\{1,\ldots,T_\epsilon\}}f(\bar{x}_t)-f(x_*),
	\\\leq&\min_{t\in\{1,\ldots,T_\epsilon\}}L\norm{\bar{x}_t-x_*},
	\\\leq&L\epsilon\sqrt{n},
	\\\leq&O(L\sqrt{n}T_\epsilon^{-\frac{1}{n}}),
\end{align}
using \autoref{def:condition}, which is convergent to $0$ for constant $L$ and $n$. Hence, the simple regret of grid search is infinitesimally small for large $T$ and Hannan consistent. This result is a consequence of the inherent hardness of the problem setting at hand. On the other hand, the average regret of grid search is given by
\begin{align}
	r_T\leq&\frac{1}{T}\sum_{t=1}^{T}f({\bar{x}}_t)-\min_{x\in\Omega}f(x),
	\\\leq& O(1),
\end{align}
i.e., in general, it does not converge (i.e., it cannot learn). Moreover, it is obvious to see that a bound on the average regret implies a bound on the simple regret. To this end, instead of the simple regret, we use the average regret for the purposes of performance evaluation and will create an algorithm that achieves $O(T^{-\frac{1}{n}})$ minimax optimal average regret. 

\section{The Algorithm}\label{sec:algorithm}
In this section, we design our algorithm, which can be used to iteratively minimize a non-convex objective function $f(\cdot)$ that satisfies the constraint in \autoref{def:condition}. Although, we do the optimization in a $n$-dimensional unit cube $[0,1]^n$, it can be straightforwardly extended for use in an arbitrarily located cube with arbitrary length after translation and scaling of the inputs. In contrast to the traditional algorithms in global optimization such as Piyavskii-Shubert variants, which work by creating lower bounding proxy functions and updating them with each new query; we utilize a fixed query construction rule to increase computational efficiency \cite{gokcesu2022low}. The framework of the algorithm is given next.

\begin{enumerate}
	\item We wrap the domain $\Omega=[0,1]^n$ into a hyper-rectangle
	\begin{align*}
		\Theta=[0,\theta^{n-1}]\times[0,\theta^{n-2}]\times\ldots\times[0,\theta^0],
	\end{align*}
	where $\theta=2^{\frac{1}{n}}$; such that we evaluate any $\tilde{x}\in\Theta$ after projecting it into $\Omega$, i.e., 
	\begin{align*}
		f(\tilde{x})=f(\argmin_{{x}\in\Omega}\norm{\tilde{x}-{x}})
	\end{align*}
	which coincides with upper-bounding its elements by $1$.
	\item At the start, we set the middle point
	\begin{align*}
		x_a=\{\theta^{-1},\theta^{-2},\ldots,\theta^{-n}\}
	\end{align*}
	and its edge vector (the distances to  the boundaries)
	\begin{align*}
		v_a=\{\theta^{-1},\theta^{-2},\ldots,\theta^{-n}\}.
	\end{align*}
	We sample $x_a$ and set its evaluation as 
	\begin{align*}
		f_a=f(\argmin_{x\in\Omega}\norm{x-x_a}).
	\end{align*}
	\item Using $x_a$, $v_a$ and $f_a$ as inputs; we determine two candidate points $x_b$, $x_c$ together with their edge vectors $v_b$, $v_c$ and their scores $s_b$, $s_c$. We add the candidates to the list of potential queries. \label{item:candidate}
	\item We sample the candidate with the lowest score $s'$ and remove it from the list. Let that query be $x'$, its evaluation be $f'=f(\argmin_{x\in\Omega}\norm{x-x'})$, and its edge vector be $v'$. \label{item:sample}
	\item We repeat Step \ref{item:candidate} with the following inputs: $x_a=x'$, $v_a=r'$, $f_a=f'$.
	\item We return to Step \ref{item:sample}.
\end{enumerate}

\begin{remark}
	After each query, the algorithm iteratively creates two new potential  queries and the parameters of the unevaluated potential query points remain unchanged.
\end{remark}

\begin{remark}
	Since after a query, two new potential queries are created, the candidate list grows linearly with the number of made queries.
\end{remark}

\begin{remark}
	Many stopping criteria can be considered including stopping after a fixed number of queries or when the queried point is sufficiently close to the optimal evaluation.
\end{remark}

\begin{remark}
	Although the algorithm is efficient in its creation of the potential queries,  the memory complexity can be improved by eliminating the queries with scores higher than the best evaluation so far (as will be apparent later).
\end{remark}


The potential query points are selected differently from the popular Piyavskii-Shubert variants. Instead of constructing proxy lower bounding functions that pass from the sampled points and minimizing them, we determine the candidate query points by splitting the hyper-rectangle implied by the queried point $x'$ together with its edge vector $v'$ among its largest dimension into two equally large hyper-rectangles. The potential new queries are the center points of these newly created hyper-rectangles. Their exact expression is given in the following.

\begin{definition}\label{thm:candidate}
	For an objective function $f(\cdot)$, given the sampled point $x_a$ and its edge vector $v_a$; the potential queries $x_b$, $x_c$ and their edge vectors $v_b$, $v_c$ are given by
	\begin{align*}
		x_b=&x_a+z,
		\\x_c=&x_a-z,
		\\v_b=&v_a-z,
		\\v_c=&v_a-z,
	\end{align*}
	where $z$ is an all zero vector except at the index $I=\argmin_{i\in\{1,2,\ldots,n\}}v_a(i)$, where $v_a(i)$ denotes the $i^{th}$ element of $v_a$. Hence,
	\begin{align*}
		z(i)=\begin{cases}
			\frac{v_a(I)}{2},& i=I
			\\0,& i\neq I
		\end{cases}.
	\end{align*}
\end{definition}

Instead of minimizing proxy lower bounding functions as in Piyavskii-Shubert variants, the scores $s_b$ and $s_c$ are created by values that completely lower bounds the respective region of the potential queries.

\begin{lemma}\label{thm:score}
	For an objective function $f(\cdot)$ that satisfies \autoref{def:condition}, when the point $x_a$ is queried with its corresponding evaluation $f_a$;  we assign the following scores $s_b$ and $s_c$ for the candidate points $x_b$, $x_c$ with their respective edge vectors $v_b$, $v_c$:
	\begin{align*}
		s_b=&f_a-L\norm{v_a},\\
		s_c=&f_a-L\norm{v_a},
	\end{align*}
	where the scores $s_b$ and $s_c$ completely lower bounds the functional evaluation of the respective regions of the queries $x_b$ and $x_c$.
	
	\begin{proof}
		A point $x_b$ with its respective edge vector $v_b$ covers the hyper-rectangle region whose center is $x_b$ where the individual distance to the boundary planes are given by its edge vector $v_b$, i.e., 
		\begin{align*}
			\mathcal{H}_b=&[x_b(1)-v_b(1),x_b(1)+v_b(1)]\\
			&\times[x_b(2)-v_b(2),x_b(2)+v_b(2)]\\
			&\times\ldots\\
			&\times[x_b(n)-v_b(n),x_b(n)+v_b(n)].
		\end{align*}
		Because of the Lipschitz continuity in \autoref{def:condition}, we have
		\begin{align}
			f(x)\leq& f_a-L\norm{x-x_a},\\
			\leq& f_a-L\norm{v_a}.
		\end{align}
		Similar arguments follow for $x_c$, $v_c$ and $\mathcal{H}_c$; hence, we reach the result.
	\end{proof}
\end{lemma}

\section{Average Regret analysis}

\begin{lemma}\label{thm:sampleRegret}
	For a given objective function $f(\cdot)$ that satisfies \autoref{def:condition}, let us sample the point $x_b$ with the respective score $s_b$ that was created after the query $x_a$ with its evaluation $f_a$. When the evaluation of the point $x_b$ is $f_b$, we have the following result.
	\begin{align}
		f_b-\min_{x\in\Omega}f(x)\leq (1+\theta)L\norm{v_b}
	\end{align} 
	where $v_b$ is the associated edge vector of the query $x_b$.
	\begin{proof}
		We know that the score of query $x_b$, i.e., $s_b$, completely lower bounds the evaluations in the hyper-rectangle $\mathcal{H}_b$ that is associated with the center point $x_b$ and the edge vector $v_b$, i.e.,
		\begin{align}
			s_b\leq \min_{x\in\mathcal{H}_b}f(x).
		\end{align}
		By the design of the algorithm, the query list consists of a set of potential queries $\mathcal{X}$ whose associated regions $\mathcal{H}$ are disjoint and their union covers up the whole search domain. Hence, we have
		\begin{align}
			s_b\leq \min_{x\in\Omega}f(x).
		\end{align} 
	Combining with \autoref{thm:score}, we have
	\begin{align}
		f_a-L\norm{v_a}\leq \min_{x\in\Omega}f(x).
	\end{align}
	Moreover, because of Lipschitz continuity from \autoref{def:condition}, we have
	\begin{align}
		f_b\leq f_a+L\norm{v_a},
	\end{align}
	where the query $x_b$ is created after the sampling of $x_a$ with its evaluation $f_a$. Thus, combining the two results, we get
	\begin{align}
		f_b\leq \min_{x\in\Omega}f(x)+(1+\theta)L\norm{v_b},
	\end{align}
	since $\theta\norm{v_b}=\norm{v_a}$ by design, which concludes the proof.
	\end{proof}
\end{lemma}




\begin{lemma}\label{thm:regret}
	The algorithm has the following cumulative regret
	\begin{align*}
		\sum_{t=1}^Tf_t-\sum_{t=1}^Tf(x_*)\leq (1+\theta)L\sum_{t=1}^T\norm{v_t}.
	\end{align*}
	for the queries points $\{x_t\}_{t=1}^T$, where $\{f_t\}_{t=1}^T$ are their evaluations and $\{v_t\}_{t=1}^T$ are their edge vectors, respectively.
	\begin{proof}
		We run the algorithm for $T$ sampling times. Let this samples be $\{x_t\}_{t=1}^T$ with their respective edge vectors $\{v_t\}_{t=1}^T$ and evaluations $\{f_t\}_{t=1}^T$. Because of Lipschitz continuity of \autoref{def:condition}, for the initial sampling, we have
		\begin{align}
			f_1-f(x_*)\leq L\norm{v_1}.
		\end{align}
		For the other samplings, i.e., $t\geq2$, we have from \autoref{thm:sampleRegret}
		\begin{align}
			f_t-f(x_*)\leq (1+\theta)L\norm{v_t}.
		\end{align}
		Thus, the cumulative regret is bounded by
		\begin{align}
			\sum_{t=1}^Tf_t-\sum_{t=1}^Tf(x_*)\leq (1+\theta)L\sum_{t=1}^T\norm{v_t},
		\end{align}
		which concludes the proof.
	\end{proof}
\end{lemma}		
\begin{theorem}\label{thm:averageregret}
	The algorithm has the following average regret bound
	\begin{align*}
		r_T\leq O(L\sqrt{n}T^{-\frac{1}{n}})
	\end{align*}
\begin{proof}
	From the construction of the algorithm, with each new sampling, the largest element in the edge vector of the sampled point is halved. Since, at the beginning of the algorithm, the initial edge vector starts as $v_1=\{\theta^{-1},\theta^{-2},\ldots,\theta^{-n}\},$ where $\theta=2^{\frac{1}{n}}$; the norm of the edge vector multiplicatively decreases by $2^{\frac{1}{n}}$. Thus, in the worst-case scenario, we will have
	\begin{align}
		\sum_{t=1}^T\norm{v_t}\leq\sum_{i=0}^{K-1}2^i\frac{V}{2^{\frac{i}{n}}}+M\frac{V}{2^{\frac{K}{n}}},
	\end{align}
	where $2^K-1+M=T$, $2^K\geq M\geq1$ and $V=\norm{v_1}$. Hence, we have
	\begin{align}
		\sum_{t=1}^T\norm{v_t}\leq&\sum_{i=0}^{K}2^i\frac{V}{2^{\frac{i}{n}}},
		\\\leq&V\sum_{i=0}^{K}2^{i\left(\frac{n-1}{n}\right)}
		\\\leq&V\frac{2^{(K+1)\frac{n-1}{n}}}{2^{\frac{n-1}{n}}-1}
		\\\leq&\frac{\theta^{n-1}}{\theta^{n-1}-1}VT^{\frac{n-1}{n}},
		\\\leq&O(\sqrt{n}T^{\frac{n-1}{n}}),
	\end{align}
	Hence, the average regret is
	\begin{align*}
		r_T\leq&O(L\sqrt{n}T^{-\frac{1}{n}}),
	\end{align*}
	which concludes the proof.
\end{proof}
\end{theorem}
		
\begin{theorem}
	The result in \autoref{thm:averageregret} is minimax optimal.
	\begin{proof}
		Suppose, we have an objective function $f(\cdot)$ whose value is globally optimal at some point $x_*$, i.e., $f(x_*)=\min_{x\in\Omega}f(x)$. Suppose, the function is such that, for some $\epsilon>0$, the $\epsilon$-neighborhood of the global optimum is linear with the Lipschitz constant $L>0$ and constant with $0$ everywhere else, i.e.,
		\begin{align}
			f(x)=\begin{cases}
				L(\norm{x-x_*}-\epsilon),& \norm{x-x_*}\leq\epsilon\\
				0,& \norm{x-x_*}>\epsilon
			\end{cases}.
		\end{align}
		Let an algorithm sample the points $\{x_t\}_{t=1}^T$. Its average regret will be given by
		\begin{align}
			r_T= \frac{1}{T}\sum_{t=1}^Tf(x_t)-f(x_*)
		\end{align}
		Suppose, $f(\cdot)$ is such that all the evaluations are valued $0$, i.e., $f(x_t)=0, \forall t$. Hence, $r_T=-f(x^*)=L\epsilon$. For this, let 
		\begin{align}
			\epsilon_*=\min_\tau\norm{x_\tau-x_*}, 
		\end{align}
		which will maximize $\epsilon$, hence, the regret. 
		Thus, the minimax regret is
		\begin{align}
			r^*_T\geq& L\min_{\{x_t\}_{t=1}^T}\max_{x_*\in\Omega}\min_\tau\norm{x_\tau-x_*}\\
			\geq&\Omega(L\sqrt{n}{T^{-\frac{1}{n}}}),
		\end{align}
		by filling the unit cube with small balls for constant $L$.
	\end{proof}
\end{theorem}
	
\bibliographystyle{ieeetran}
\bibliography{double_bib}

\begin{thebibliography}{10}
\providecommand{\url}[1]{#1}
\csname url@samestyle\endcsname
\providecommand{\newblock}{\relax}
\providecommand{\bibinfo}[2]{#2}
\providecommand{\BIBentrySTDinterwordspacing}{\spaceskip=0pt\relax}
\providecommand{\BIBentryALTinterwordstretchfactor}{4}
\providecommand{\BIBentryALTinterwordspacing}{\spaceskip=\fontdimen2\font plus
\BIBentryALTinterwordstretchfactor\fontdimen3\font minus
  \fontdimen4\font\relax}
\providecommand{\BIBforeignlanguage}[2]{{%
\expandafter\ifx\csname l@#1\endcsname\relax
\typeout{** WARNING: IEEEtran.bst: No hyphenation pattern has been}%
\typeout{** loaded for the language `#1'. Using the pattern for}%
\typeout{** the default language instead.}%
\else
\language=\csname l@#1\endcsname
\fi
#2}}
\providecommand{\BIBdecl}{\relax}
\BIBdecl

\bibitem{pinter1991global}
J.~D. Pint{\'e}r, ``Global optimization in action,'' \emph{Scientific
  American}, vol. 264, pp. 54--63, 1991.

\bibitem{cesa_book}
N.~Cesa-Bianchi and G.~Lugosi, \emph{Prediction, learning, and games}.\hskip
  1em plus 0.5em minus 0.4em\relax Cambridge university press, 2006.

\bibitem{poor_book}
H.~V. Poor, \emph{An Introduction to Signal Detection and Estimation}.\hskip
  1em plus 0.5em minus 0.4em\relax NJ: Springer, 1994.

\bibitem{rios2013derivative}
L.~M. Rios and N.~V. Sahinidis, ``Derivative-free optimization: a review of
  algorithms and comparison of software implementations,'' \emph{Journal of
  Global Optimization}, vol.~56, no.~3, pp. 1247--1293, 2013.

\bibitem{jones1998efficient}
D.~R. Jones, M.~Schonlau, and W.~J. Welch, ``Efficient global optimization of
  expensive black-box functions,'' \emph{Journal of Global optimization},
  vol.~13, no.~4, pp. 455--492, 1998.

\bibitem{jain2017non}
P.~Jain and P.~Kar, ``Non-convex optimization for machine learning,''
  \emph{Foundations and Trends{\textregistered} in Machine Learning}, vol.~10,
  no. 3-4, pp. 142--336, 2017.

\bibitem{hansen1991number}
P.~Hansen, B.~Jaumard, and S.-H. Lu, ``On the number of iterations of
  piyavskii's global optimization algorithm,'' \emph{Mathematics of Operations
  Research}, vol.~16, no.~2, pp. 334--350, 1991.

\bibitem{basso1982iterative}
P.~Basso, ``Iterative methods for the localization of the global maximum,''
  \emph{SIAM Journal on Numerical Analysis}, vol.~19, no.~4, pp. 781--792,
  1982.

\bibitem{brochu2010tutorial}
E.~Brochu, V.~M. Cora, and N.~De~Freitas, ``A tutorial on bayesian optimization
  of expensive cost functions, with application to active user modeling and
  hierarchical reinforcement learning,'' \emph{arXiv preprint arXiv:1012.2599},
  2010.

\bibitem{boyd2004convex}
S.~Boyd, S.~P. Boyd, and L.~Vandenberghe, \emph{Convex optimization}.\hskip 1em
  plus 0.5em minus 0.4em\relax Cambridge university press, 2004.

\bibitem{nesterov2003introductory}
Y.~Nesterov, \emph{Introductory lectures on convex optimization: A basic
  course}.\hskip 1em plus 0.5em minus 0.4em\relax Springer Science \& Business
  Media, 2003, vol.~87.

\bibitem{bubeck2015convex}
S.~Bubeck, ``Convex optimization: Algorithms and complexity,''
  \emph{Foundations and Trends{\textregistered} in Machine Learning}, vol.~8,
  no. 3-4, pp. 231--357, 2015.

\bibitem{munos2014bandits}
R.~Munos, ``From bandits to monte-carlo tree search: The optimistic principle
  applied to optimization and planning,'' \emph{Foundations and Trends® in
  Machine Learning}, vol.~7, no.~1, pp. 1--129, 2014.

\bibitem{shalev2012online}
S.~Shalev-Shwartz \emph{et~al.}, ``Online learning and online convex
  optimization,'' \emph{Foundations and Trends{\textregistered} in Machine
  Learning}, vol.~4, no.~2, pp. 107--194, 2012.

\bibitem{spall2005introduction}
J.~C. Spall, \emph{Introduction to stochastic search and optimization:
  estimation, simulation, and control}.\hskip 1em plus 0.5em minus 0.4em\relax
  John Wiley \& Sons, 2005, vol.~65.

\bibitem{gokcesu2018density}
K.~Gokcesu and S.~S. Kozat, ``Online density estimation of nonstationary
  sources using exponential family of distributions,'' \emph{{IEEE} Trans.
  Neural Networks Learn. Syst.}, vol.~29, no.~9, pp. 4473--4478, 2018.

\bibitem{willems}
F.~M.~J. Willems, ``Coding for a binary independent
  piecewise-identically-distributed source.'' \emph{IEEE Transactions on
  Information Theory}, vol.~42, no.~6, pp. 2210--2217, 1996.

\bibitem{gokcesu2018anomaly}
K.~Gokcesu and S.~S. Kozat, ``Online anomaly detection with minimax optimal
  density estimation in nonstationary environments,'' \emph{{IEEE} Trans.
  Signal Process.}, vol.~66, no.~5, pp. 1213--1227, 2018.

\bibitem{coding2}
G.~I. Shamir and N.~Merhav, ``Low-complexity sequential lossless coding for
  piecewise-stationary memoryless sources,'' \emph{IEEE Transactions on
  Information Theory}, vol.~45, no.~5, pp. 1498--1519, Jul 1999.

\bibitem{neyshabouri2018asymptotically}
M.~M. Neyshabouri, K.~Gokcesu, H.~Gokcesu, H.~Ozkan, and S.~S. Kozat,
  ``Asymptotically optimal contextual bandit algorithm using hierarchical
  structures,'' \emph{IEEE transactions on neural networks and learning
  systems}, vol.~30, no.~3, pp. 923--937, 2018.

\bibitem{cesa-bianchi}
S.~Bubeck and N.~Cesa{-}Bianchi, ``Regret analysis of stochastic and
  nonstochastic multi-armed bandit problems,'' \emph{Foundations and Trends in
  Machine Learning}, vol.~5, no.~1, pp. 1--122, 2012.

\bibitem{gokcesu2018bandit}
K.~{Gokcesu} and S.~S. {Kozat}, ``An online minimax optimal algorithm for
  adversarial multiarmed bandit problem,'' \emph{IEEE Transactions on Neural
  Networks and Learning Systems}, vol.~29, no.~11, pp. 5565--5580, 2018.

\bibitem{tnnls3}
H.~R. Berenji and P.~Khedkar, ``Learning and tuning fuzzy logic controllers
  through reinforcements,'' \emph{IEEE Transactions on Neural Networks},
  vol.~3, no.~5, pp. 724--740, Sep 1992.

\bibitem{ozkan}
H.~Ozkan, M.~A. Donmez, S.~Tunc, and S.~S. Kozat, ``A deterministic analysis of
  an online convex mixture of experts algorithm,'' \emph{IEEE Transactions on
  Neural Networks and Learning Systems}, vol.~26, no.~7, pp. 1575--1580, July
  2015.

\bibitem{tnnls1}
R.~Song, F.~L. Lewis, and Q.~Wei, ``Off-policy integral reinforcement learning
  method to solve nonlinear continuous-time multiplayer nonzero-sum games,''
  \emph{IEEE Transactions on Neural Networks and Learning Systems}, vol.~PP,
  no.~99, pp. 1--10, 2016.

\bibitem{gokcesu2016prediction}
N.~D. Vanli, K.~Gokcesu, M.~O. Sayin, H.~Yildiz, and S.~S. Kozat, ``Sequential
  prediction over hierarchical structures,'' \emph{IEEE Transactions on Signal
  Processing}, vol.~64, no.~23, pp. 6284--6298, Dec 2016.

\bibitem{singer}
A.~C. Singer and M.~Feder, ``Universal linear prediction by model order
  weighting,'' \emph{IEEE Transactions on Signal Processing}, vol.~47, no.~10,
  pp. 2685--2699, Oct 1999.

\bibitem{tnnls4}
J.~Moody and M.~Saffell, ``Learning to trade via direct reinforcement,''
  \emph{IEEE Transactions on Neural Networks}, vol.~12, no.~4, pp. 875--889,
  Jul 2001.

\bibitem{gokcesu2019outlier}
K.~Gokcesu, M.~M. Neyshabouri, H.~Gokcesu, and S.~S. Kozat, ``Sequential
  outlier detection based on incremental decision trees,'' \emph{{IEEE} Trans.
  Signal Process.}, vol.~67, no.~4, pp. 993--1005, 2019.

\bibitem{gokcesu2016nested}
I.~Delibalta, K.~Gokcesu, M.~Simsek, L.~Baruh, and S.~S. Kozat, ``Online
  anomaly detection with nested trees,'' \emph{{IEEE} Signal Process. Lett.},
  vol.~23, no.~12, pp. 1867--1871, 2016.

\bibitem{gokcesu2017online}
K.~Gokcesu and S.~S. Kozat, ``Online anomaly detection with minimax optimal
  density estimation in nonstationary environments,'' \emph{IEEE Transactions
  on Signal Processing}, vol.~66, no.~5, pp. 1213--1227, 2017.

\bibitem{bartlett2019simple}
P.~L. Bartlett, V.~Gabillon, and M.~Valko, ``A simple parameter-free and
  adaptive approach to optimization under a minimal local smoothness
  assumption,'' in \emph{Algorithmic Learning Theory}.\hskip 1em plus 0.5em
  minus 0.4em\relax PMLR, 2019, pp. 184--206.

\bibitem{grill2015black}
J.-B. Grill, M.~Valko, and R.~Munos, ``Black-box optimization of noisy
  functions with unknown smoothness,'' \emph{Advances in Neural Information
  Processing Systems}, vol.~28, pp. 667--675, 2015.

\bibitem{piyavskii1972algorithm}
S.~Piyavskii, ``An algorithm for finding the absolute extremum of a function,''
  \emph{USSR Computational Mathematics and Mathematical Physics}, vol.~12,
  no.~4, pp. 57--67, 1972.

\bibitem{shubert1972sequential}
B.~O. Shubert, ``A sequential method seeking the global maximum of a
  function,'' \emph{SIAM Journal on Numerical Analysis}, vol.~9, no.~3, pp.
  379--388, 1972.

\bibitem{schoen1982sequential}
F.~Schoen, ``On a sequential search strategy in global optimization problems,''
  \emph{Calcolo}, vol.~19, no.~3, pp. 321--334, 1982.

\bibitem{mayne1984outer}
D.~Q. Mayne and E.~Polak, ``Outer approximation algorithm for nondifferentiable
  optimization problems,'' \emph{Journal of Optimization Theory and
  Applications}, vol.~42, no.~1, pp. 19--30, 1984.

\bibitem{mladineo1986algorithm}
R.~H. Mladineo, ``An algorithm for finding the global maximum of a multimodal,
  multivariate function,'' \emph{Mathematical Programming}, vol.~34, no.~2, pp.
  188--200, 1986.

\bibitem{shen1987interval}
Z.~Shen and Y.~Zhu, ``An interval version of shubert's iterative method for the
  localization of the global maximum,'' \emph{Computing}, vol.~38, no.~3, pp.
  275--280, 1987.

\bibitem{horst1987convergence}
R.~Horst and H.~Tuy, ``On the convergence of global methods in multiextremal
  optimization,'' \emph{Journal of Optimization Theory and Applications},
  vol.~54, no.~2, pp. 253--271, 1987.

\bibitem{hansen1995lipschitz}
P.~Hansen and B.~Jaumard, ``Lipschitz optimization,'' in \emph{Handbook of
  global optimization}.\hskip 1em plus 0.5em minus 0.4em\relax Springer, 1995,
  pp. 407--493.

\bibitem{horst2013global}
R.~Horst and H.~Tuy, \emph{Global optimization: Deterministic
  approaches}.\hskip 1em plus 0.5em minus 0.4em\relax Springer Science \&
  Business Media, 2013.

\bibitem{breiman1993deterministic}
L.~Breiman and A.~Cutler, ``A deterministic algorithm for global
  optimization,'' \emph{Mathematical Programming}, vol.~58, no.~1, pp.
  179--199, 1993.

\bibitem{baritompa1994accelerations}
W.~Baritompa and A.~Cutler, ``Accelerations for global optimization covering
  methods using second derivatives,'' \emph{Journal of Global Optimization},
  vol.~4, no.~3, pp. 329--341, 1994.

\bibitem{sergeyev1998global}
Y.~D. Sergeyev, ``Global one-dimensional optimization using smooth auxiliary
  functions,'' \emph{Mathematical Programming}, vol.~81, no.~1, pp. 127--146,
  1998.

\bibitem{ellaia2012modified}
R.~Ellaia, M.~Z. Es-Sadek, and H.~Kasbioui, ``Modified piyavskii’s global
  one-dimensional optimization of a differentiable function,'' \emph{Applied
  Mathematics}, vol.~3, pp. 1306--1320, 2012.

\bibitem{brent2013algorithms}
R.~P. Brent, \emph{Algorithms for minimization without derivatives}.\hskip 1em
  plus 0.5em minus 0.4em\relax Courier Corporation, 2013.

\bibitem{gokcesu2021regret}
K.~Gokcesu and H.~Gokcesu, ``Regret analysis of global optimization in
  univariate functions with lipschitz derivatives,'' \emph{arXiv preprint
  arXiv:2108.10859}, 2021.

\bibitem{danilin1971estimation}
Y.~M. Danilin, ``Estimation of the efficiency of an absolute-minimum-finding
  algorithm,'' \emph{USSR Computational Mathematics and Mathematical Physics},
  vol.~11, no.~4, pp. 261--267, 1971.

\bibitem{malherbe2017global}
C.~Malherbe and N.~Vayatis, ``Global optimization of lipschitz functions,'' in
  \emph{International Conference on Machine Learning}.\hskip 1em plus 0.5em
  minus 0.4em\relax PMLR, 2017, pp. 2314--2323.

\bibitem{bouttier2020regret}
C.~Bouttier, T.~Cesari, and S.~Gerchinovitz, ``Regret analysis of the
  piyavskii-shubert algorithm for global lipschitz optimization,'' \emph{arXiv
  preprint arXiv:2002.02390}, 2020.

\bibitem{gokcesu2022low}
K.~Gokcesu and H.~Gokcesu, ``Low regret binary sampling method for efficient
  global optimization of univariate functions,'' \emph{arXiv preprint
  arXiv:2201.07164}, 2022.

\end{thebibliography}
\end{document}